\documentclass[10pt,twocolumn,letterpaper]{article}

\usepackage{wacv}              

\usepackage{graphicx}
\usepackage{amsmath}
\usepackage{amssymb}
\usepackage{booktabs}
\usepackage{multirow}
\usepackage[table,xcdraw]{xcolor}
\usepackage[normalem]{ulem}
\useunder{\uline}{\ul}{}

\usepackage[pagebackref,breaklinks,colorlinks]{hyperref}

\usepackage[capitalize]{cleveref}
\crefname{section}{Sec.}{Secs.}
\Crefname{section}{Section}{Sections}
\Crefname{table}{Table}{Tables}
\crefname{table}{Tab.}{Tabs.}

\def\wacvPaperID{470} 
\def\confName{WACV}
\def\confYear{2024}

\begin{document}

\title{Outline-Guided Object Inpainting with Diffusion Models}

\author{Markus Pobitzer\\
ETH Zürich\\
{\tt\small pobmarku@ethz.ch}
\and
Filip Janicki\\
IBM Research Zurich\\
{\tt\small fja@zurich.ibm.com}
\and
Mattia Rigotti\\
IBM Research Zurich\\
{\tt\small MRG@zurich.ibm.com}
\and
Cristiano Malossi\\
IBM Research Zurich\\
{\tt\small acm@zurich.ibm.com}
}
\maketitle

\begin{abstract}
    Instance segmentation datasets play a crucial role in training accurate and robust computer vision models.
    However, obtaining accurate mask annotations to produce high-quality segmentation datasets is a costly and labor-intensive process. 
    In this work, we show how this issue can be mitigated by starting with small annotated instance segmentation datasets and augmenting them 
    to effectively obtain a sizeable annotated dataset.
    We achieve that by creating variations of the available annotated object instances in a way that preserves the provided mask annotations, thereby resulting in new image-mask pairs to be added to the set of annotated images.
    Specifically, we generate new images using a diffusion-based inpainting model to fill out the masked area with a desired object class by guiding the diffusion through the object outline.
    We show that the object outline provides a simple, but also reliable and convenient training-free guidance signal for the underlying inpainting model that is often sufficient to fill out the mask with an object of the correct class without further text guidance and preserve the correspondence between generated images and the mask annotations with high precision.
    Our experimental results reveal that our method successfully generates realistic variations of object instances, preserving their shape characteristics while introducing diversity within the augmented area.
    We also show that the proposed method can naturally be combined with text guidance and other image augmentation techniques.
\end{abstract}

\begin{figure*}
\begin{center}
\includegraphics[width=\textwidth]{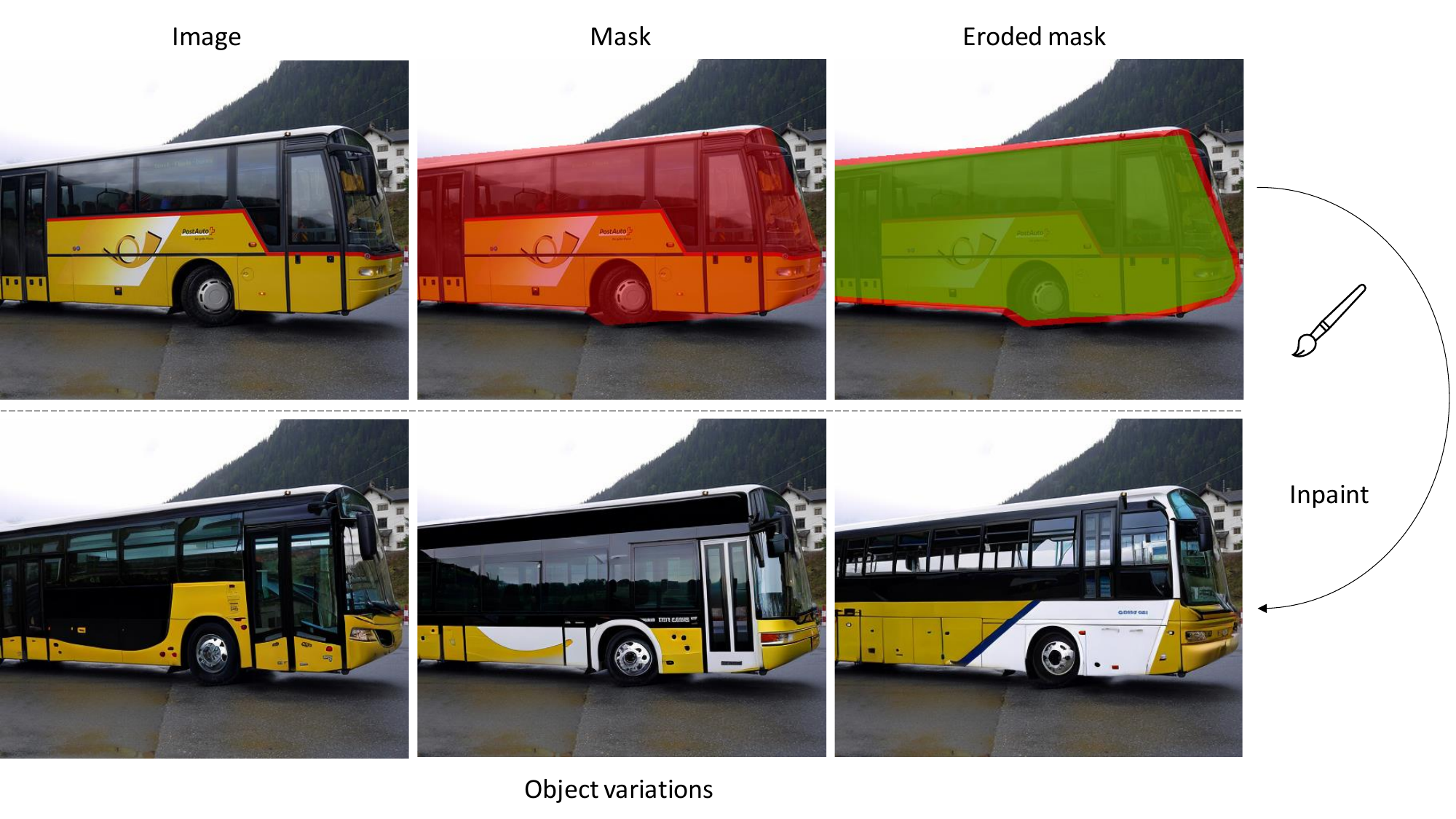}
\end{center}
   \caption{Overview of our proposed method. Given an image, a mask for an object, and the associated class, we create various variations of the object through inpainting. For this, we erode the original mask to give the inpainting model guidance through the outline of the object. Through this, it creates a variation of the original object and is less likely to inpaint the background. Additionally, we can provide text guidance, such as the associated class. The method is not limited to any specific type of object and works well in challenging scenes. It can easily be combined with existing image augmentations.}
\label{fig:teaser}
\end{figure*}

\section{Introduction}
\label{sec:intro}

Semantic Instance Segmentation is the task of detecting objects in an image and creating a pixel-wise mask for each.
Computer vision algorithms based on machine learning have drastically increased the accuracy of this task \cite{he2017mask, liu2021swin, dosovitskiy2020image}.
However, such algorithms need large training datasets, which are typically costly to produce as the process of providing pixel-wise annotations for all objects in a given image is time- and labor-intensive.

With the rise of generative image models, such as Latent Diffusion \cite{rombach2022high} a new way of creating synthetic training data became possible.
In \cite{azizi2023synthetic} the authors have shown that generated images can help image classification.
A recent work \cite{karazija23diffusion} showed images generated by diffusion models contain important information that can be used for zero-shot open vocabulary segmentation.

In our work, we take a closer look at how diffusion models can be used for a more classic image augmentation, specifically in the instance segmentation scenario. 
The main challenge towards this goal is to augment images and the corresponding mask annotations such that they will remain consistent with each other.
One possible strategy is to generate new images and let existing segmentation models such as SAM \cite{kirillov2023segment} generate masks that would be used as ground truth annotation.
This has the downside that the segmentation model can introduce inaccuracies or wrong annotations.

A different approach initiated on the hills of success of diffusion models is to synthesize new images using these models while conditioning the generation by guiding it with an existing ground truth mask (see the Related Work section).
In our work, we show that existing inpainting models \cite{rombach2022high} already achieve a similar result if we guide them indirectly with a much simpler mechanism consisting of simply providing an object outline to be filled out by the diffusion model.
This method is motivated by our observations on the impressive capability of diffusion models to generate the continuation of partial objects, in addition to their already known capability to outpaint backgrounds.
Taking advantage of this property and applying it to object outlines allows us to side-step the notorious difficulty inherent in guiding inpainting through text alignment.
In addition, our method is completely training-free, which means that it can leverage available pre-trained diffusion models out of the box.


We demonstrate how our method can be used to augment already existing instance segmentation datasets such as COCO \cite{lin2014microsoft}. COCO contains image mask pairs of 80 different object classes commonly found in real life.
With the help of the ground truth masks in the dataset, we inpaint the objects with stable diffusion to create new variations of the same object class.
As mentioned, as guidance we use the outline of the object (by eroding the mask) and the class of the object in the form of text guidance.

In summary, this is the overview of our contribution:
\begin{enumerate}
    \itemsep0em
  \item we propose an approach for creating synthetic variations of objects in an instance segmentation dataset that closely follows the outline of the original object preserving the original object mask;
  \item we show that object outlines provide a simple indirect, but robust guidance mechanism for latent diffusion inpainting;
  \item we demonstrate the adaptability of this guidance mechanism under several image operations, such as resizing, flipping, and color shifts, and demonstrate its use for data augmentation.
\end{enumerate}

\section{Related Work}
We review diffusion models (DMs) for image generation and their adaptability to different applications. The authors of \cite{sohl2015deep, ho2020denoising} have introduced how Diffusion Probabilistic Models can be used to generate images. Previously, mainly Generative adversarial networks (GANs) \cite{goodfellow2020generative} such as \cite{karras2020analyzing} had been used to synthesize realistic-looking images. Continuing the work of \cite{ho2020denoising} we have seen rapid development of DMs in the last few years \cite{song2020denoising, nichol2021improved, dhariwal2021diffusion, ho2022classifier} leading up to Imagen \cite{saharia2022photorealistic}, Latent Diffusion \cite{rombach2022high}, and its popular implementation Stable Diffusion (SD).

\subsection{Latent Diffusion Models}
Latent diffusion models were introduced in \cite{rombach2022high}, and decisively contributed to the popularity of diffusion models by reducing their computational requirements for the generation of high-resolution images.
Their proposed solution was to apply the diffusion process in a lower dimensional latent space and train an encoder/decoder that can map from image to latent space. 
Stable Diffusion (SD) is a latent diffusion model trained on Laion-5B \cite{schuhmann2022laion} a dataset containing image-text pairs. Through the vast knowledge contained in Laion-5B, the model is able to create realistic-looking content. The content of the images can be guided through text with the help of classifier free guidance \cite{ho2022classifier}. Other guidance methodologies on top of SD were introduced in ControlNet \cite{zhang2023adding} and \cite{mou2023t2i}. Similar to \cite{huang2023composer}, they incorporate guidance through edges \cite{canny1986computational}, poses, and many more. These guidance methods work for general diffusion models too,
but generally have the disadvantage that they have to be used while training the diffusion model, and cannot therefore be combined with fixed models that were pretrained without consideration for the guidance mechanism.

\subsection{Inpainting}
Inpainting is the task of filling out parts of an image specified by a mask. Where some other convolution methods fail, Lama \cite{suvorov2022resolution} can inpaint challenging parts of an image thanks to a global and local understanding of the scene.
In our work, we mainly focus on inpainting with diffusion models. Already \cite{sohl2015deep} showed the possibilities of inpainting with Diffusion models and RePaint \cite{lugmayr2022repaint} was able to achieve realistic results even when using extreme masks by altering the reverse diffusion process. However, RePaint was mainly developed for pixel space and not latent space as used in SD and, even though it can still be applied in the latent space of SD, pixel-level control is lost, resulting in inaccurate mask-object correspondence.

Other works such as GLIDE \cite{nichol2021glide} and Blended diffusion \cite{avrahami2022blended} showed that text-guided image inpaintings can work well with diffusion models. For SD, fine-tuning additional parameters helped to create a well-working inpainting method while using the trained SD model \cite{rombach2022high} as a starting point.
SmartBrush \cite{xie2023smartbrush} has shown that with additional regularization it is possible to fine-tune a diffusion model so that it can fill out the complete masked area with the inpainted object while being guided with text and mask -- something standard SD inpainting fails at.

A related GAN based approach was introduced in \cite{zeng2022shape}. Instead of using the object outline, the authors propose to guess the object class by the shape of the mask.

\begin{figure}[!ht]
\begin{center}
\includegraphics[width=\columnwidth]{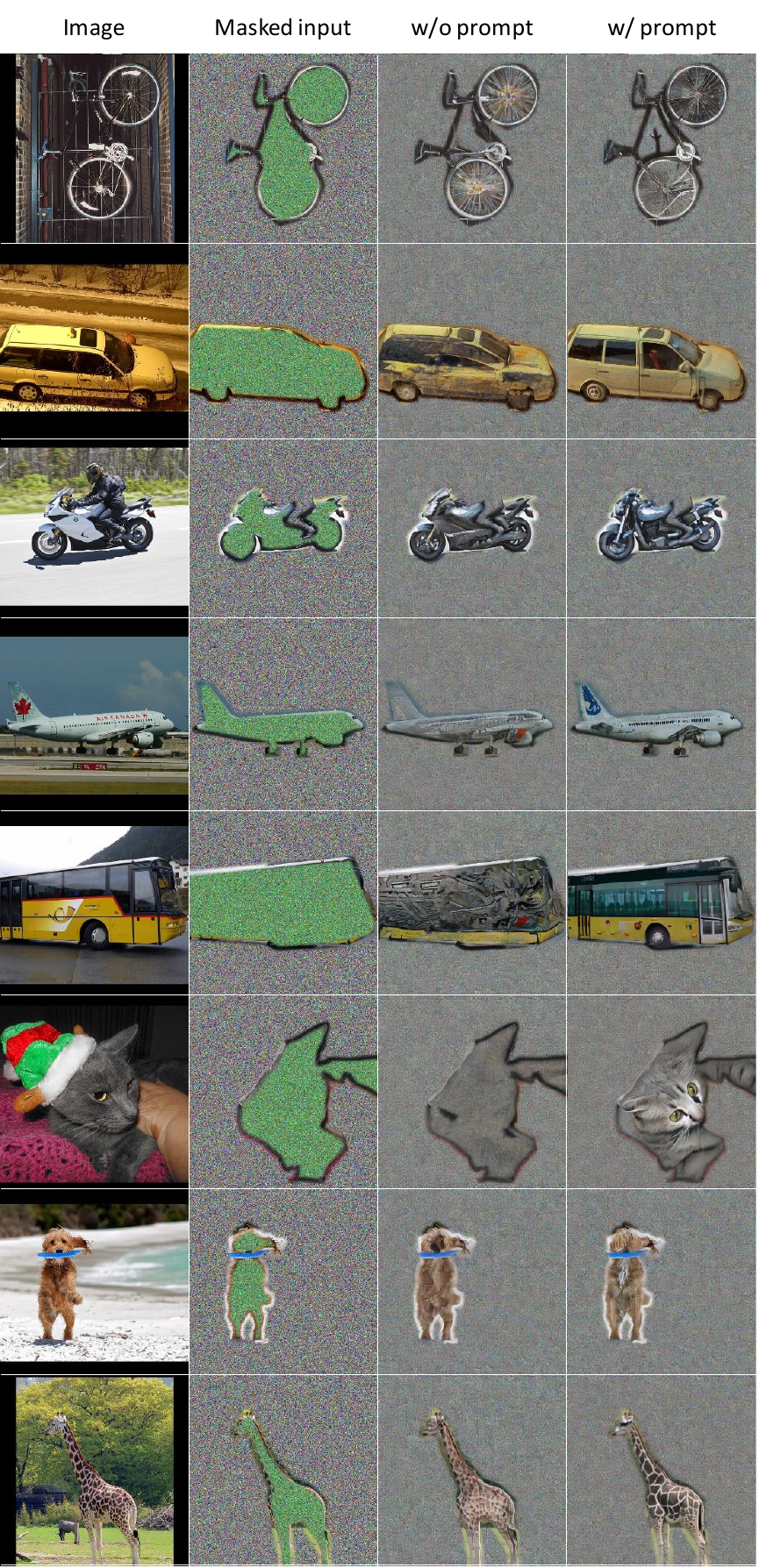}

\end{center}
   \caption{Inpaintings without and with text guidance where only the outline of the object is provided. The first column shows the original images, and the second column is the actual input, where the masked area to inpaint is highlighted in green. The background was replaced with noise to remove possible object connections with the scene and focus only on the outline. The last two columns show inpainting results without a prompt and with the object class as the prompt.}
\label{fig:outline}
\end{figure}

\section{Method}

\begin{figure*}
\begin{center}
\includegraphics[width=\textwidth]{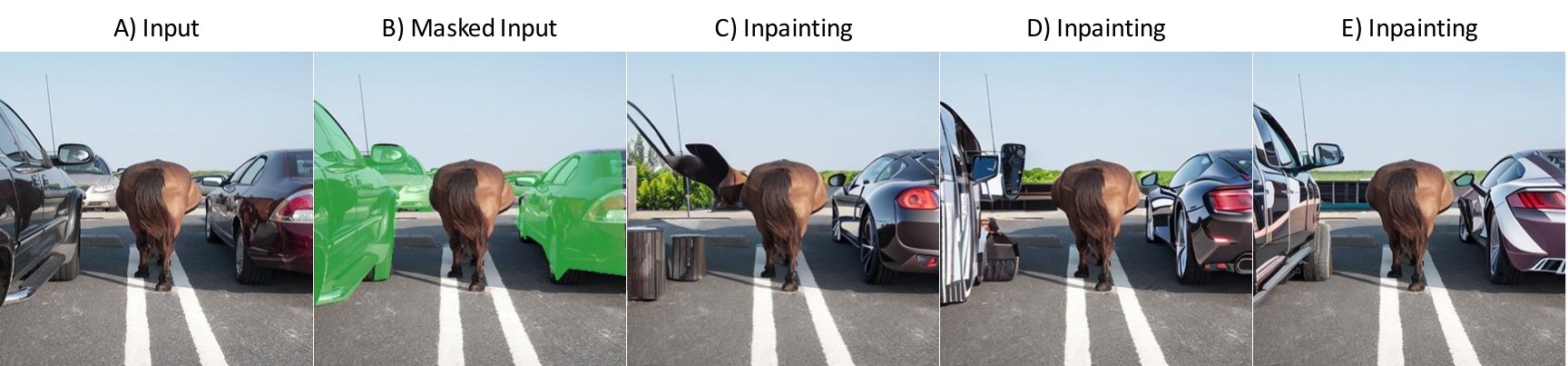}

\end{center}
   \caption{Here we show an example A) that failed when we tried to remove the cars with the original COCO masks depicted in B). The text prompt ``photograph of a beautiful empty scene, highest quality settings'' was used and follows \cite{rombach2022high} for background inpainting. The negative prompt contained ``car''. In the inpaintings C), D), and E) we see that a car has been inpainted on the right of the horse. We argue this is the case since the mask does not completely cover the car and the model uses the outline of the object as guidance.}
\label{fig:object_removal_problem}
\end{figure*}

In this section, we show how an instance segmentation dataset can be augmented with diffusion models.
We demonstrate our method using the inpainting model trained on top of SD \footnote{https://huggingface.co/stabilityai/stable-diffusion-2-inpainting}.
When we refer to an inpainting model, we mean this model if not otherwise specified. The model was trained with large masks similar to \cite{suvorov2022resolution}. Therefore, it can fill out big patches of an image in a realistic manner. We use the Diffusers \cite{von-platen-etal-2022-diffusers} library to perform inference. For our dataset we use COCO \cite{lin2014microsoft} an image dataset containing common objects, their instances, segmentations, and captions. We mainly rely on the object segmentations and the corresponding object class.

When running initial experiments on removing objects from a scene given their mask annotations, we noticed that the inpainting model failed if the mask did not completely overlap with the object, see \cref{fig:object_removal_problem}.
Various approaches to mitigate this problem \cite{gandikota2023erasing} did not give satisfactory results.
We argue that this phenomenon stems from the way the inpainting model was trained, namely by filling out patches of an image, thereby endowing it with the capacity to naturally extend objects.

This observation motivates our proposal to use mask annotations to partially remove objects by morphological erosion of the bulk of the objects, leaving their outline intact.
We then use the inpainting model to fill out the object, relying on its capacity to latch onto the remaining outline to guide the generation.
This simultaneously ensures that the newly generated image is consistent with the overall context in the scene and that the inpainted object remains still within the mask annotation of the original image, which can therefore serve as a high-quality ground truth annotation of the newly generated image.
Additionally, we guide it through text with the information of the object class. An overview of our procedure is presented in \cref{fig:teaser}.

Our chosen text prompt is ``Photo of a $\mathit{C^*}$'' or ``Photo of several $\mathit{C^*}$'' (for inpainting more than one instance), where $\mathit{C^*}$ is the class of the object. We chose this prompt by evaluating several options on Laion-5B through clip retrieval \cite{beaumont-2022-clip-retrieval} since this is the dataset that was used to train the model. We chose ``Photo'' over ``Image'', ``Picture'', and ``Photograph'' since it indicates a real scene.
We found ``Photograph'' is often linked with a depiction of a picture frame too, something we do not want.
We also experimented with the use of a negative prompt (see supplementary material). Whenever a text prompt helps to guide the output toward the object class, the negative prompt should help to steer the output away from generations that do not look photo-realistic. This prompt is used in all experiments where the use of a negative prompt is mentioned.

For the COCO dataset we set a conservative erosion kernel of 12x12 pixels, with this we can reasonably well ensure that the outline is not over-eroded even in the case of complex scene outlines and inaccurate masks.
On the other hand, we lose some variation freedom through the relatively thick outline and lose the opportunity to apply the augmentation for very small objects.

As discussed in \cite{zhu2023designing} the encoding/decoding step of the latent space in SD is lossy, which causes alterations also to parts of an image that should stay unchanged.
To circumvent this problem we keep the original image in the unmodified part and blend the edges of the two parts with a Gaussian filter.

\section{Experiments}

\textbf{Few Shot Dataset}

COCO-20i is a subset of the COCO dataset and is extensively used for few shot instance segmentation \cite{shaban2017one, nguyen2019feature}. The 80 object classes get split up into four independent sets, called fold 0 to fold 3. Each fold therefore contains 20 classes used for the few shot task. From each class in the fold, we sample five images, called 5-shots, to train an instance segmentation model. This leaves us with 100 training images for an experiment, so-called support set. More details about COCO-20i can be found in \cite{shaban2017one}.

For our experiments we used a Mask R-CNN \cite{he2017mask} model with a ResNet \cite{he2016deep} as the backbone. We train it from scratch with the Detectron2 \cite{wu2019detectron2} framework and use their provided training script for the model R50-FPN. For each fold, we pre-train the model with the 60 unrelated classes for 130k iterations, a batch size of 64, and a learning rate of $0.1$. In the following experiments, an NVIDIA A100 40GB was used.

The first experimental setup entails sampling five images (5-shot) out of the 20 related classes, creating a total of 100 real images that are used as the baseline, called the support set. Five such support sets are created per fold, and then each gets augmented with a certain factor of generated images. The average precision (AP) as defined in \cite{lin2014microsoft} was used as the evaluation metric. Results can be seen in \cref{tab:coco20i_metric}, where a shown entry is the average of the five experiments in a fold and the train set indicates the total number of training images. For the training step of the experiments, we started with the pre-trained model on the unrelated classes, trained only on the set containing images of the 20 related classes, the batch size was reduced to 16 and a learning rate of $0.025$ was used.

\begin{table}
    \centering
    \begin{tabular}{l|cccc|c}
    \toprule
    {}&     \multicolumn{4}{c|}{AP COCO-20i Fold}& {} \\
    {Train Set}&     0 &      1 &      2 &      3& {AVG} \\
    \midrule
    100 & 6.57 & 10.21 & 10.65 & 10.53 & 9.490 \\
    \midrule
    120 & 6.61 & 10.02 & 10.81 & 10.73 & +0.053\\
    140 & 6.76 & 10.33 & 11.02 & \underline{10.99} & +0.283\\
    160 & 6.78 & 10.41 & 10.98 & 10.93 & +0.284\\
    200 & \underline{6.94} & 10.58 & 11.21 & 10.85 & \underline{+0.406}\\
    300 & 6.74 & \underline{10.63} & 11.20 & 10.84 & +0.363\\
    400 & 6.80 & 10.58 & \underline{11.25} & 10.91 & +0.394\\
    \bottomrule
    \end{tabular}
    \caption{Scaling the few-shot training dataset by adding augmented images. The mean segmentation AP over all folds for the baseline is 9.49. The indicated improvements show the results obtained with the addition of generated data.}
    \label{tab:coco20i_metric}
\end{table}

The second experimental setup uses a fixed one-to-one ratio between generated images and real ones, in total 100 real images as the baseline and ontop adding 100 generated ones for comparison. We test the minimal area of the object, meaning that we select only images that contain objects with an area bigger equals the minimal area. Our assumption is that the method works better with bigger objects. The values of the minimal areas follow the object sizes defined in COCO where small objects have an area smaller than $32^2$ pixels, medium-sized objects are smaller than $96^2$ pixels, and large objects are at least $96^2$ pixels. To better evaluate the size of the erosion kernel, we select kernel sizes of $0, 6$, and $12$ pixels. Additionally, we look at the impact of using the text prompt alone (positive) and adding the negative prompt (negative). The results can be seen in \cref{tab:fold_comparison}

\begin{table*}
    \centering
    \begin{tabular}{c|c|c|c|c|c|c||c|c}
    \toprule
    \textbf{Min Area} &
      \textbf{Erosion} &
      \textbf{Prompt} &
      \multicolumn{4}{c||}{\textbf{Fold}} &
      \textbf{AVG} &
      \textbf{Diff.} \\
                \hline
                 &                      &                      & 0          & 1           & 2           & 3           &              &                                     \\
                 \hline\hline
                 & {} & {} & 6.34       & 10.13       & 10.52       & 9.83        & 9.207        & \cellcolor[HTML]{FFFFFF}0.000       \\
                 \hline
                 & 0                    & positive             & 6.30       & 9.84        & 10.10       & 9.75        & 8.998        & \cellcolor[HTML]{F87173}-0.209      \\
                 & 0                    & negative             & 6.36       & 9.95        & 10.09       & 9.55        & 8.985        & \cellcolor[HTML]{F8696B}-0.222      \\
                 & 6                    & positive             & 6.66       & 10.38       & 10.63       & 10.14       & 9.454        & \cellcolor[HTML]{A2D8B0}0.248       \\
                 & 6                    & negative             & {\ul 6.75} & 10.34       & 10.42       & 10.05       & 9.390        & \cellcolor[HTML]{BAE3C5}0.183       \\
                 & 12                   & positive             & 6.46       & 10.39       & {\ul 10.86} & 10.22       & 9.482        & \cellcolor[HTML]{97D4A7}0.275       \\
        \multirow{-7}{*}{$\geq 0^2$}                     & 12 & negative & 6.55 & {\ul 10.44} & 10.67 & {\ul 10.35} & {\ul 9.501} & \cellcolor[HTML]{90D1A1}0.295 \\
        \hline\hline
                 & {} & {} & 6.69       & 10.88       & 10.80       & 11.71       & 10.020       & \cellcolor[HTML]{FFFFFF}0.000       \\
                 \hline
                 & 0                    & positive             & 6.65       & 10.87       & 10.85       & 11.33       & 9.928        & \cellcolor[HTML]{FCC0C1}-0.092      \\
                 & 0                    & negative             & 6.63       & 11.04       & 10.91       & 11.36       & 9.985        & \cellcolor[HTML]{FDE7E7}-0.035      \\
                 & 6                    & positive             & {\ul 7.02} & 11.21       & 11.21       & 11.96       & 10.349       & \cellcolor[HTML]{82CB96}0.329       \\
                 & 6                    & negative             & 6.92       & 11.13       & 11.21       & 11.89       & 10.288       & \cellcolor[HTML]{9AD5A9}0.268       \\
                 & 12                   & positive             & 7.02       & {\ul 11.27} & {\ul 11.23} & {\ul 11.98} & {\ul 10.374} & \cellcolor[HTML]{79C88E}{0.354} \\
        \multirow{-7}{*}{$\geq 32^2$} & 12 & negative & 6.91 & 11.21       & 11.21 & 11.82       & 10.287      & \cellcolor[HTML]{9AD5AA}0.267 \\
        \hline\hline
                 & {} & {} & 6.08       & 10.41       & 10.09       & 10.41       & 9.247        & \cellcolor[HTML]{FFFFFF}0.000       \\
                 \hline
                 & 0                    & positive             & 5.96       & 10.00       & 9.96        & 10.25       & 9.041        & \cellcolor[HTML]{F87375}-0.206      \\
                 & 0                    & negative             & 5.94       & 10.03       & 10.03       & 10.23       & 9.056        & \cellcolor[HTML]{F87D7F}-0.191      \\
                 & 6                    & positive             & {\ul 6.39} & 10.67       & 10.55       & 10.69       & 9.576        & \cellcolor[HTML]{83CB96}0.329       \\
                 & 6                    & negative             & 6.31       & 10.71       & 10.54       & 10.68       & 9.561        & \cellcolor[HTML]{88CE9B}0.314       \\
                 & 12                   & positive             & 6.38       & 10.74       & {\ul 10.62} & {\ul 10.89} & {\ul 9.658}  & \cellcolor[HTML]{63BE7B}{0.411} \\
        \multirow{-7}{*}{$\geq 96^2$} & 12 & negative & 6.31 & {\ul 10.75} & 10.54 & 10.86       & 9.616       & \cellcolor[HTML]{73C589}0.369 \\
        \bottomrule
    \end{tabular}
    \caption{Ablation study on evaluating segmentation AP when using a minimum area for the inpainted objects, different sized erosion kernels, and usage of a negative prompt. The first row of each min area represents the baseline. The difference is taken to the average value for the baseline.}
    \label{tab:fold_comparison}
\end{table*}

For the final experiment, we select objects with a min area of $96^2$ pixels, use a kernel of size 12 by 12 pixels for erosion, and the positive text prompt as guidance. To better show the deviation of the results we sampled ten image sets for each fold (compared to five as before), each containing five images per class (5-shot), and rerun each experiment several times such that an estimation of the standard deviation is possible. Results can be seen in \cref{tab:stddev_comparison}.

\begin{table*}
\centering
\begin{tabular}{l|cccc||c}
\toprule
\multicolumn{1}{c|}{} & \multicolumn{4}{c||}{AP COCO-20i Fold}                                & \multicolumn{1}{c}{} \\
                     & 0              & 1              & 2               & 3               &  {AVG}                       \\
                     \hline \hline
baseline             & 6.01 (± 0.112) & 10.51 (± 0.16) & 9.91 (± 0.139)  & 10.64 (± 0.113) & 9.27 (± 0.132)          \\
ours                 & 6.27 (± 0.108) & 10.93 (± 0.17) & 10.52 (± 0.162) & 11.03 (± 0.141) & 9.69 (± 0.147) \\ 
\bottomrule

\end{tabular}
\caption{Comparison of segmentation AP with standard deviation in brackets. Only images containing large objects were considered, an erosion kernel of size 12, and a prompt containing the object class was used.}
    \label{tab:stddev_comparison}
\end{table*}
 
\textbf{FID Evaluation}. We follow SmartBrush in evaluating our method using the Fréchet inception distance (FID) \cite{frechet1957distance, heusel2017gans}.
To make the results comparable we resize each image in the test set of COCO to a size of 512 by 512. Then we sample two masks for each image, and erode them with a kernel of size 12. 
We have noticed that occasionally some masks vanish after erosion; in those cases, we revert back to using the original masks. The final results get resized to 256 by 256 as in SmartBrush.
This yields 9311 images that can be used for data augmentation through inpainting the eroded regions. The FID score was calculated between the original COCO images and the inpainted ones.

We also report the ``Local FID'' score, comparing only the cutout of the bounding box around the object to focus more on the inpainted region. To make the results comparable we report the scores of the direct output from the inpainting method without the overlay of the original image on top of the unmasked part. Results can be seen in \cref{tab:fid}.

\textbf{CLIP Score}. Clip score was introduced in \cite{radford2021learning} and has been shown to capture text-image alignment.
With the help of the provided COCO captions, we apply the evaluation method on the previously inpainted images. Results are shown in \cref{tab:fid}.

\begin{table*}[!ht]
    \centering
    \begin{tabular}{lrrr}
    \hline
        ~ & FID $\downarrow$ & Local FID $\downarrow$ & CLIP Score $\uparrow$ \\ \hline
        Blended Diffusion \cite{avrahami2022blended} & 8.16 & 26.25 & 0.244\\
        GLIDE \cite{nichol2021glide} & 6.98 & 24.25 & 0.235\\
        SD Inpainting (v1.5) \cite{rombach2022high} & 6.54 & 15.16 & 0.243\\
        SmartBrush \cite{xie2023smartbrush} & 5.76 & 9.80 & 0.249 \\ 
        SD Inpainting (v2.0) \cite{rombach2022high} & 5.24 & 9.16 & 0.297 \\ \hline
        Ours & {\ul 4.38} & {\ul 7.14} & {\ul 0.299}\\
    \end{tabular}
    \caption{Results of the FID metric and Clip Score. Evaluation procedure and some results from \cite{xie2023smartbrush}.}
    \label{tab:fid}
\end{table*}

\begin{figure}
\begin{center}
\includegraphics[width=\columnwidth]{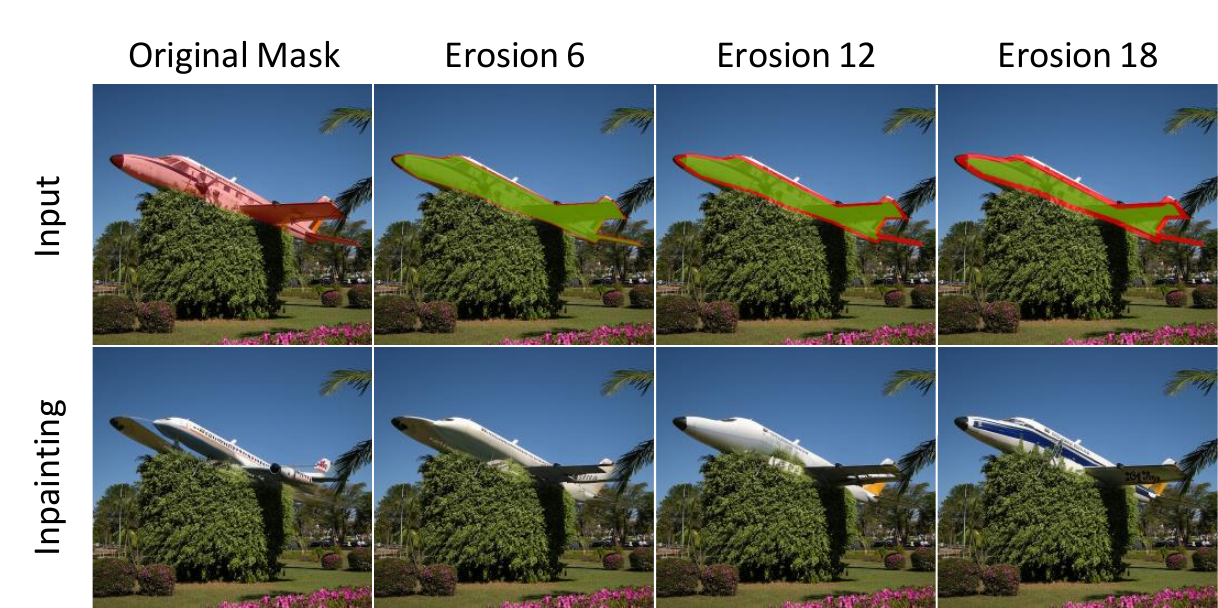}

\end{center}
   \caption{Here we show the impact of the erosion size. We start with the original mask (erosion 0) and then gradually increase the erosion kernel in steps of 6 pixels. The top row shows the image with the applied eroded mask. The bottom row shows the corresponding generated outputs. If the erosion size is too small the inpainting model does not completely follow the outline of the airplane.}
\label{fig:erosion_size}
\end{figure}

\begin{figure}
\begin{center}
\includegraphics[width=\columnwidth]{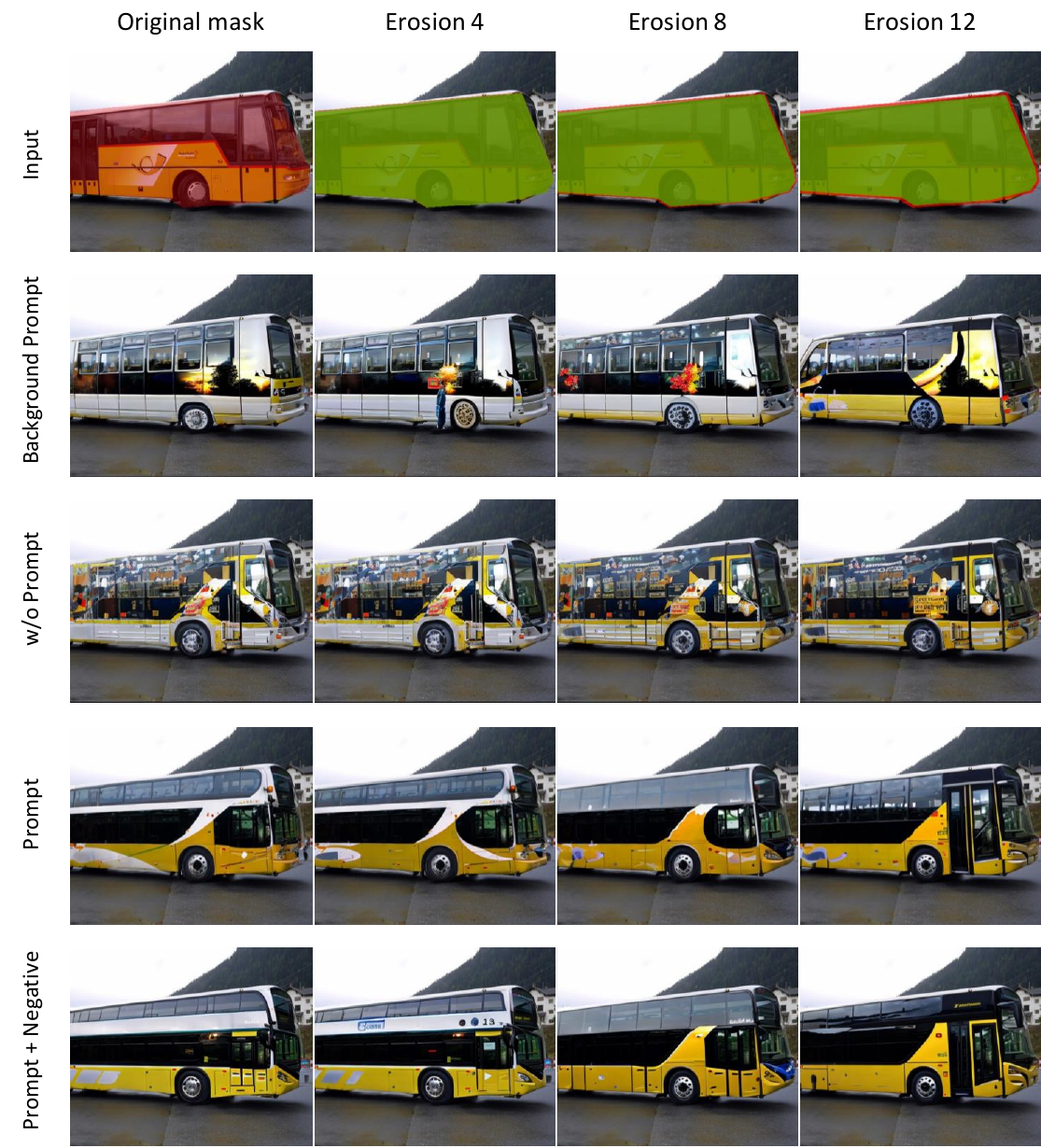}

\end{center}
   \caption{Comparison of different prompts with different erosion sizes on an image with class label ``bus''. In the second row, we have the background prompt: ``photograph of a beautiful empty scene, highest quality settings'' but the output still resembles a bus. This illustrates the text misalignment issue mentioned in \cite{xie2023smartbrush} but in a more general sense. The third row is generated with no prompt and the model still coherently inpaints a bus. This is also the case for the original mask since it did not cover the whole bus. Row 4 has our standard prompt and row 5 adds the negative prompt.}
\label{fig:prompt}
\end{figure}

\subsection{Outline and Text Guidance}
For the rest of our evaluations, we use the COCO dataset and apply padding if needed and cropping afterward to get to a resolution of 512 by 512, since this is the size on which the inpainting model was trained.

In \cref{fig:outline} we show the impact of the object outline.
For this, we remove the rest of the scene, except for the outline, by replacing it with noise. This leaves only the outline of the object to guide the inpainting.

To better understand the indirect guidance through the object outline we show results of using different erosion kernels in \cref{fig:erosion_size} and \cref{fig:prompt}.
Guidance through object outline can also be seen in \cref{fig:object_removal_problem}.
The outline guidance provides the inpainting model with the exact location where the object should be in the scene. For our use cases, the mask should mainly be filled with the desired object.

An evaluation of the importance of guidance through text input can be seen in \cref{fig:prompt}, where we evaluate several different text prompts, including the effect of the negative prompts. The text prompt should help better specify which features of a certain class should be inpainted. It may be ambiguous to create the desired object just by its outline and therefore the text prompt can be used to provide more specific control.

\section{Evaluation}
\textbf{Few Shot Dataset}
In \cref{tab:coco20i_metric} the evaluation of scaling the number of generated images compared to a fixed set of real images showed that the best performance was achieved around a one-to-one ratio. Similar to the findings in \cite{azizi2023synthetic} we did see a decrease in performance when drastically increasing the number of generated images.

Continuing with these findings, we structured the following experiments such that we keep the one-to-one ratio. In \cref{tab:fold_comparison} we found a tendency that an increase in the size of the object leads to better performance of the method. At the same time, we noticed that the inpainting operation without eroding the mask did not lead to an improvement, whereas a kernel size of 12 achieved the best results from the tested sizes. From the shown results we did not notice an improvement when using the additional negative prompt.

For the final experiment in \cref{tab:stddev_comparison} the previously best settings were used to show the standard deviation for a more reliable comparison to the baseline. An improvement over the baseline can be noted.

\begin{figure*}
\begin{center}
\includegraphics[width=\textwidth]{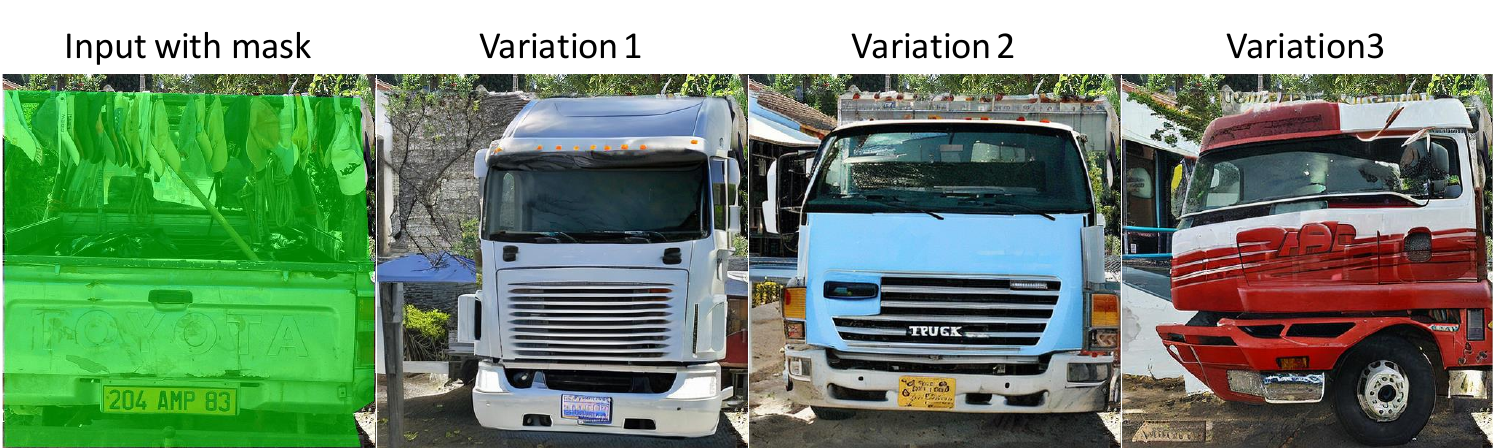}

\end{center}
   \caption{An example where the method fails. The class ``truck'' does not fill the indicated area in the original image. We used an erosion of 6 and ``truck'' as the prompt. No negative prompt was used.}
\label{fig:failure_1}
\end{figure*}

\textbf{FID Evaluation}. In \cref{tab:fid} we see that our method outperforms the current state-of-the-art approaches, in the sense that it generates samples that are perceptually closer to the original dataset.
This is quantified both in terms of FID score over the whole image and as well as when computed only within the bounding box of the object. Our method achieves good results since it can replicate some important visual features, such as color and pattern, of the original image with relative ease.
At this point, we should note that our method in the comparison in \cref{tab:fid} takes advantage of the fact that we are using the actual outline of the object, a strong additional guidance, but more importantly a set of pixels that contains information about the visual feature of the original object.
As an example, the model can fairly easily estimate the color of the inpainted object just by its outline, compare outputs in \cref{fig:prompt} where the bus variations have yellow parts similar to the input. Additionally, our method does not change small objects since the eroded mask becomes very small.

\textbf{Clip Score}. The Clip score is an indication of how well the text prompt aligns with the image. In \cref{tab:fid} we have the comparison to the other methods. Also here, we see that our method achieves very good results, indicating the inpainted parts closely follow the content of the original image.

\subsection{Object Outline Guidance}

The outline of an object is a very strong guidance since it contains the size and often also indicates the orientation of the object.
Additionally, it can provide other features such as color, shape, and texture. As we can see in \cref{fig:outline} the outline alone can be a strong guidance for the inpainting method.
Not only does it allow to continue the seen texture but it also provides a means to infer the object class, such as in the first example where a bicycle gets inpainted without text guidance. 
We however also observe cases where the object outline is not informative enough to infer the object class. 
A hypothesis for why the method fails is that the shape is too generic (in the bus case) or does not align well with the object class (as in the image with the cat wearing a hat). In such cases, the addition of text guidance does a good job of steering the generation in the right direction.

From our experiments with different degrees of mask erosion, we conclude that the object outline is the crucial element allowing our method to completely inpaint the masked area with the desired object.
As seen in \cref{fig:erosion_size}, the text prompt alone typically leads to only partially filled-out masks in the absence of a large enough outline of the original object. This also gets confirmed in \cref{tab:fold_comparison}, where inpainting the original object leads to a decrease in performance, and only using an eroded mask increases performance.

Finally, we observe that the object outline can also limit the diversity of the produced image variations. 
This effect again strongly depends on the size of the erosion kernel, as it trades off prior information about the object and freedom in the generation.
In the supplementary material, a showcase of the method, and visual examples, of how the method can be combined with additional prepossessing steps to the original image to create more diverse outputs.

\subsection{Text guidance}

We next studied the effect of the text prompt and found that ``Image of $\mathit{C^*}$'' or just ``$\mathit{C^*}$'' as prompt also gave good results.
\cref{fig:outline} shows that text guidance can help in challenging situations to better align the output with the desired class.
To better gauge the role of the prompts, we show the effect of fixing the input and varying the prompt in \cref{fig:prompt}. The original mask of the bus is not well aligned with the object and we notice that even without a prompt or one that should inpaint an empty background, visual features of a bus appear. The object outline has more importance in guiding the output however, the prompt helps in producing outputs that align well with real-life objects.



\subsection{Shortcomings}
\cref{fig:failure_1} provides a failure case, where the masked area is not completely filled by the object.
For better visualization, we used a smaller erosion size of 6 in this example.
This type of gailers seems to mainly arise when the mask fills a large part of the image or the object is placed in an unnatural environment, as seen in the example in \cref{fig:erosion_size}.
In such cases, to guide the inpainting model towards better compliance of the boundaries of the mask, the erosion size can be increased. The use of the negative prompt also had a positive effect.

During the evaluation of the method on the few-shot dataset, we also noticed that some COCO classes did not benefit from the augmentation. Namely the parking meter and frisbee, wherein the first one the model had problems inpainting fine details and text, and with the frisbee class it added strange color artifacts on the object. More details can be found in the supplementary material.

Finally, we noticed failure cases of our method corresponding to cases where the objects were clearly out of distribution.
As an example, objects that are rotated in an unnatural way (e.g. upside-down vehicles) tend to be filled out in their natural position.

\section{Conclusion}

In this work, we propose a simple data augmentation pipeline for instance segmentation datasets consisting of images-mask annotation pairs.
Our method takes advantage of a straightforward, but robust conditioning mechanism for diffusion-based inpainting model, namely the fact that providing partial information on a visual object corresponding to its outline conditions the inpainting to reliably produce high-quality variations of the original object that maintains the object boundary, thereby remaining consistent with the mask annotation paired with the original image.
Our experiments show that the object outline is a strong factor in guiding image generation. In a few shot setting, it can effectively augment images to improve segmentation AP. However, the relationship between the scene, background, object outline, and mask shape to the image generation is not completely clear. These could be interesting topics for future work.


{\small
\bibliographystyle{ieee_fullname}
\bibliography{egbib}

\begin{thebibliography}{10}\itemsep=-1pt

\bibitem{avrahami2022blended}
Omri Avrahami, Dani Lischinski, and Ohad Fried.
\newblock Blended diffusion for text-driven editing of natural images.
\newblock In {\em Proceedings of the IEEE/CVF Conference on Computer Vision and Pattern Recognition}, pages 18208--18218, 2022.

\bibitem{azizi2023synthetic}
Shekoofeh Azizi, Simon Kornblith, Chitwan Saharia, Mohammad Norouzi, and David~J Fleet.
\newblock Synthetic data from diffusion models improves imagenet classification.
\newblock {\em arXiv preprint arXiv:2304.08466}, 2023.

\bibitem{beaumont-2022-clip-retrieval}
Romain Beaumont.
\newblock Clip retrieval: Easily compute clip embeddings and build a clip retrieval system with them.
\newblock \url{https://github.com/rom1504/clip-retrieval}, 2022.

\bibitem{canny1986computational}
John Canny.
\newblock A computational approach to edge detection.
\newblock {\em IEEE Transactions on pattern analysis and machine intelligence}, pages 679--698, 1986.

\bibitem{dhariwal2021diffusion}
Prafulla Dhariwal and Alexander Nichol.
\newblock Diffusion models beat gans on image synthesis.
\newblock {\em Advances in Neural Information Processing Systems}, 34:8780--8794, 2021.

\bibitem{dosovitskiy2020image}
Alexey Dosovitskiy, Lucas Beyer, Alexander Kolesnikov, Dirk Weissenborn, Xiaohua Zhai, Thomas Unterthiner, Mostafa Dehghani, Matthias Minderer, Georg Heigold, Sylvain Gelly, et~al.
\newblock An image is worth 16x16 words: Transformers for image recognition at scale.
\newblock {\em arXiv preprint arXiv:2010.11929}, 2020.

\bibitem{frechet1957distance}
Maurice Fr{\'e}chet.
\newblock Sur la distance de deux lois de probabilit{\'e}.
\newblock In {\em Annales de l'ISUP}, volume~6, pages 183--198, 1957.

\bibitem{gandikota2023erasing}
Rohit Gandikota, Joanna Materzynska, Jaden Fiotto-Kaufman, and David Bau.
\newblock Erasing concepts from diffusion models, 2023.

\bibitem{goodfellow2020generative}
Ian Goodfellow, Jean Pouget-Abadie, Mehdi Mirza, Bing Xu, David Warde-Farley, Sherjil Ozair, Aaron Courville, and Yoshua Bengio.
\newblock Generative adversarial networks.
\newblock {\em Communications of the ACM}, 63(11):139--144, 2020.

\bibitem{he2017mask}
Kaiming He, Georgia Gkioxari, Piotr Doll{\'a}r, and Ross Girshick.
\newblock Mask r-cnn.
\newblock In {\em Proceedings of the IEEE international conference on computer vision}, pages 2961--2969, 2017.

\bibitem{he2016deep}
Kaiming He, Xiangyu Zhang, Shaoqing Ren, and Jian Sun.
\newblock Deep residual learning for image recognition.
\newblock In {\em Proceedings of the IEEE conference on computer vision and pattern recognition}, pages 770--778, 2016.

\bibitem{heusel2017gans}
Martin Heusel, Hubert Ramsauer, Thomas Unterthiner, Bernhard Nessler, and Sepp Hochreiter.
\newblock Gans trained by a two time-scale update rule converge to a local nash equilibrium.
\newblock {\em Advances in neural information processing systems}, 30, 2017.

\bibitem{ho2020denoising}
Jonathan Ho, Ajay Jain, and Pieter Abbeel.
\newblock Denoising diffusion probabilistic models.
\newblock {\em Advances in Neural Information Processing Systems}, 33:6840--6851, 2020.

\bibitem{ho2022classifier}
Jonathan Ho and Tim Salimans.
\newblock Classifier-free diffusion guidance.
\newblock {\em arXiv preprint arXiv:2207.12598}, 2022.

\bibitem{huang2023composer}
Lianghua Huang, Di Chen, Yu Liu, Yujun Shen, Deli Zhao, and Jingren Zhou.
\newblock Composer: Creative and controllable image synthesis with composable conditions.
\newblock {\em arXiv preprint arXiv:2302.09778}, 2023.

\bibitem{karazija23diffusion}
Laurynas Karazija, Iro Laina, Andrea Vedaldi, and Christian Rupprecht.
\newblock {D}iffusion {M}odels for {Z}ero-{S}hot {O}pen-{V}ocabulary {S}egmentation.
\newblock {\em arXiv preprint}, 2023.

\bibitem{karras2020analyzing}
Tero Karras, Samuli Laine, Miika Aittala, Janne Hellsten, Jaakko Lehtinen, and Timo Aila.
\newblock Analyzing and improving the image quality of stylegan.
\newblock In {\em Proceedings of the IEEE/CVF conference on computer vision and pattern recognition}, pages 8110--8119, 2020.

\bibitem{kirillov2023segment}
Alexander Kirillov, Eric Mintun, Nikhila Ravi, Hanzi Mao, Chloe Rolland, Laura Gustafson, Tete Xiao, Spencer Whitehead, Alexander~C Berg, Wan-Yen Lo, et~al.
\newblock Segment anything.
\newblock {\em arXiv preprint arXiv:2304.02643}, 2023.

\bibitem{lin2014microsoft}
Tsung-Yi Lin, Michael Maire, Serge Belongie, James Hays, Pietro Perona, Deva Ramanan, Piotr Doll{\'a}r, and C~Lawrence Zitnick.
\newblock Microsoft coco: Common objects in context.
\newblock In {\em Computer Vision--ECCV 2014: 13th European Conference, Zurich, Switzerland, September 6-12, 2014, Proceedings, Part V 13}, pages 740--755. Springer, 2014.

\bibitem{liu2021swin}
Ze Liu, Yutong Lin, Yue Cao, Han Hu, Yixuan Wei, Zheng Zhang, Stephen Lin, and Baining Guo.
\newblock Swin transformer: Hierarchical vision transformer using shifted windows.
\newblock In {\em Proceedings of the IEEE/CVF international conference on computer vision}, pages 10012--10022, 2021.

\bibitem{lugmayr2022repaint}
Andreas Lugmayr, Martin Danelljan, Andres Romero, Fisher Yu, Radu Timofte, and Luc Van~Gool.
\newblock Repaint: Inpainting using denoising diffusion probabilistic models.
\newblock In {\em Proceedings of the IEEE/CVF Conference on Computer Vision and Pattern Recognition}, pages 11461--11471, 2022.

\bibitem{mou2023t2i}
Chong Mou, Xintao Wang, Liangbin Xie, Jian Zhang, Zhongang Qi, Ying Shan, and Xiaohu Qie.
\newblock T2i-adapter: Learning adapters to dig out more controllable ability for text-to-image diffusion models.
\newblock {\em arXiv preprint arXiv:2302.08453}, 2023.

\bibitem{nguyen2019feature}
Khoi Nguyen and Sinisa Todorovic.
\newblock Feature weighting and boosting for few-shot segmentation.
\newblock In {\em Proceedings of the IEEE/CVF International Conference on Computer Vision}, pages 622--631, 2019.

\bibitem{nichol2021glide}
Alex Nichol, Prafulla Dhariwal, Aditya Ramesh, Pranav Shyam, Pamela Mishkin, Bob McGrew, Ilya Sutskever, and Mark Chen.
\newblock Glide: Towards photorealistic image generation and editing with text-guided diffusion models.
\newblock {\em arXiv preprint arXiv:2112.10741}, 2021.

\bibitem{nichol2021improved}
Alexander~Quinn Nichol and Prafulla Dhariwal.
\newblock Improved denoising diffusion probabilistic models.
\newblock In {\em International Conference on Machine Learning}, pages 8162--8171. PMLR, 2021.

\bibitem{radford2021learning}
Alec Radford, Jong~Wook Kim, Chris Hallacy, Aditya Ramesh, Gabriel Goh, Sandhini Agarwal, Girish Sastry, Amanda Askell, Pamela Mishkin, Jack Clark, et~al.
\newblock Learning transferable visual models from natural language supervision.
\newblock In {\em International conference on machine learning}, pages 8748--8763. PMLR, 2021.

\bibitem{rombach2022high}
Robin Rombach, Andreas Blattmann, Dominik Lorenz, Patrick Esser, and Bj{\"o}rn Ommer.
\newblock High-resolution image synthesis with latent diffusion models.
\newblock In {\em Proceedings of the IEEE/CVF Conference on Computer Vision and Pattern Recognition}, pages 10684--10695, 2022.

\bibitem{saharia2022photorealistic}
Chitwan Saharia, William Chan, Saurabh Saxena, Lala Li, Jay Whang, Emily~L Denton, Kamyar Ghasemipour, Raphael Gontijo~Lopes, Burcu Karagol~Ayan, Tim Salimans, et~al.
\newblock Photorealistic text-to-image diffusion models with deep language understanding.
\newblock {\em Advances in Neural Information Processing Systems}, 35:36479--36494, 2022.

\bibitem{schuhmann2022laion}
Christoph Schuhmann, Romain Beaumont, Richard Vencu, Cade Gordon, Ross Wightman, Mehdi Cherti, Theo Coombes, Aarush Katta, Clayton Mullis, Mitchell Wortsman, et~al.
\newblock Laion-5b: An open large-scale dataset for training next generation image-text models.
\newblock {\em arXiv preprint arXiv:2210.08402}, 2022.

\bibitem{shaban2017one}
Amirreza Shaban, Shray Bansal, Zhen Liu, Irfan Essa, and Byron Boots.
\newblock One-shot learning for semantic segmentation.
\newblock {\em arXiv preprint arXiv:1709.03410}, 2017.

\bibitem{sohl2015deep}
Jascha Sohl-Dickstein, Eric Weiss, Niru Maheswaranathan, and Surya Ganguli.
\newblock Deep unsupervised learning using nonequilibrium thermodynamics.
\newblock In {\em International Conference on Machine Learning}, pages 2256--2265. PMLR, 2015.

\bibitem{song2020denoising}
Jiaming Song, Chenlin Meng, and Stefano Ermon.
\newblock Denoising diffusion implicit models.
\newblock {\em arXiv preprint arXiv:2010.02502}, 2020.

\bibitem{suvorov2022resolution}
Roman Suvorov, Elizaveta Logacheva, Anton Mashikhin, Anastasia Remizova, Arsenii Ashukha, Aleksei Silvestrov, Naejin Kong, Harshith Goka, Kiwoong Park, and Victor Lempitsky.
\newblock Resolution-robust large mask inpainting with fourier convolutions.
\newblock In {\em Proceedings of the IEEE/CVF winter conference on applications of computer vision}, pages 2149--2159, 2022.

\bibitem{von-platen-etal-2022-diffusers}
Patrick von Platen, Suraj Patil, Anton Lozhkov, Pedro Cuenca, Nathan Lambert, Kashif Rasul, Mishig Davaadorj, and Thomas Wolf.
\newblock Diffusers: State-of-the-art diffusion models.
\newblock \url{https://github.com/huggingface/diffusers}, 2022.

\bibitem{wu2019detectron2}
Yuxin Wu, Alexander Kirillov, Francisco Massa, Wan-Yen Lo, and Ross Girshick.
\newblock Detectron2.
\newblock \url{https://github.com/facebookresearch/detectron2}, 2019.

\bibitem{xie2023smartbrush}
Shaoan Xie, Zhifei Zhang, Zhe Lin, Tobias Hinz, and Kun Zhang.
\newblock Smartbrush: Text and shape guided object inpainting with diffusion model.
\newblock In {\em Proceedings of the IEEE/CVF Conference on Computer Vision and Pattern Recognition}, pages 22428--22437, 2023.

\bibitem{zeng2022shape}
Yu Zeng, Zhe Lin, and Vishal~M Patel.
\newblock Shape-guided object inpainting.
\newblock {\em arXiv preprint arXiv:2204.07845}, 2022.

\bibitem{zhang2023adding}
Lvmin Zhang and Maneesh Agrawala.
\newblock Adding conditional control to text-to-image diffusion models.
\newblock {\em arXiv preprint arXiv:2302.05543}, 2023.

\bibitem{zhu2023designing}
Zixin Zhu, Xuelu Feng, Dongdong Chen, Jianmin Bao, Le Wang, Yinpeng Chen, Lu Yuan, and Gang Hua.
\newblock Designing a better asymmetric vqgan for stablediffusion.
\newblock {\em arXiv preprint arXiv:2306.04632}, 2023.

\end{thebibliography}
}

\end{document}


\title{Supplementary Material}


\maketitle


\section{Method showcase}
\begin{figure}
\begin{center}
\includegraphics[width=\columnwidth]{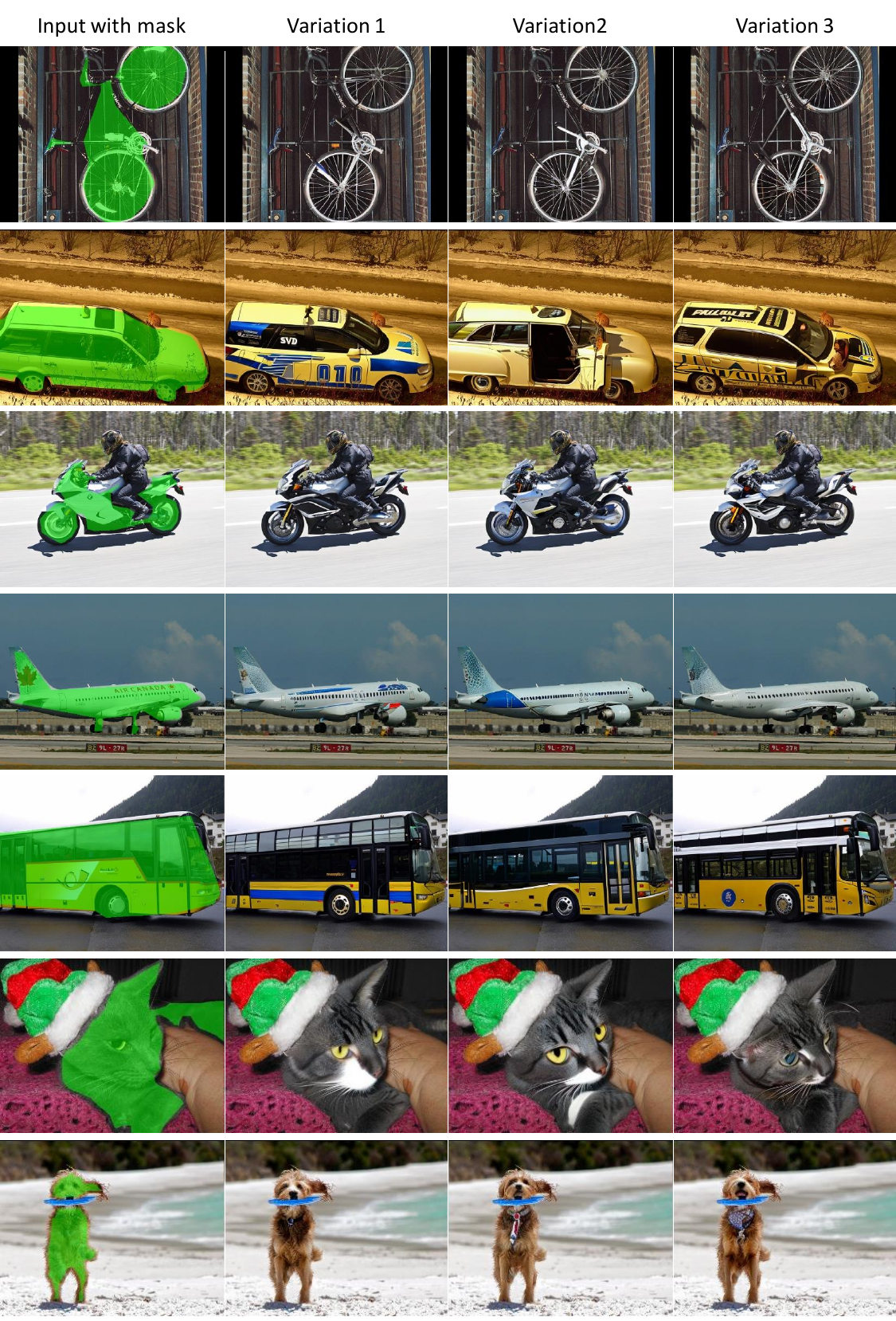}

\end{center}
   \caption{Results of our method applied to a subset of COCO classes. We used an erosion kernel of size 12, the prompt based on the class, and the mentioned negative prompt. We selected the examples to show the versatility of the method and its generalization.}
\label{fig:showcase}
\end{figure}

Figure \ref{fig:showcase} showcases several examples of all the discussed elements in the main paper coming together and producing high-quality and diverse variations of the original segmented image.

\section{Negative text prompt}
The used negative text prompt: ``disfigured, kitsch, ugly, oversaturated, grain, low-res, Deformed, blurry, bad anatomy, disfigured, poorly drawn face, mutation, mutated, extra limb, ugly, poorly drawn hands, missing limb, blurry, floating limbs, disconnected limbs, malformed hands, long neck, long body, ugly, disgusting, poorly drawn, childish, mutilated, mangled, surreal''. The experiments did not show an improvement when using it, other text may be more suitable for the prompt but this was not further investigated.

\section{Data Augmentations on input image}
Our method can naturally be combined with data augmentations on the input image and mask. Thanks to the strong image understanding of the underlying inpainting model, image color augmentations such as gray-scale, image inverting and many more are followed naturally. We used albumentations  as our main library to do the augmentations.

\begin{figure*}
\begin{center}
\includegraphics[width=\textwidth]{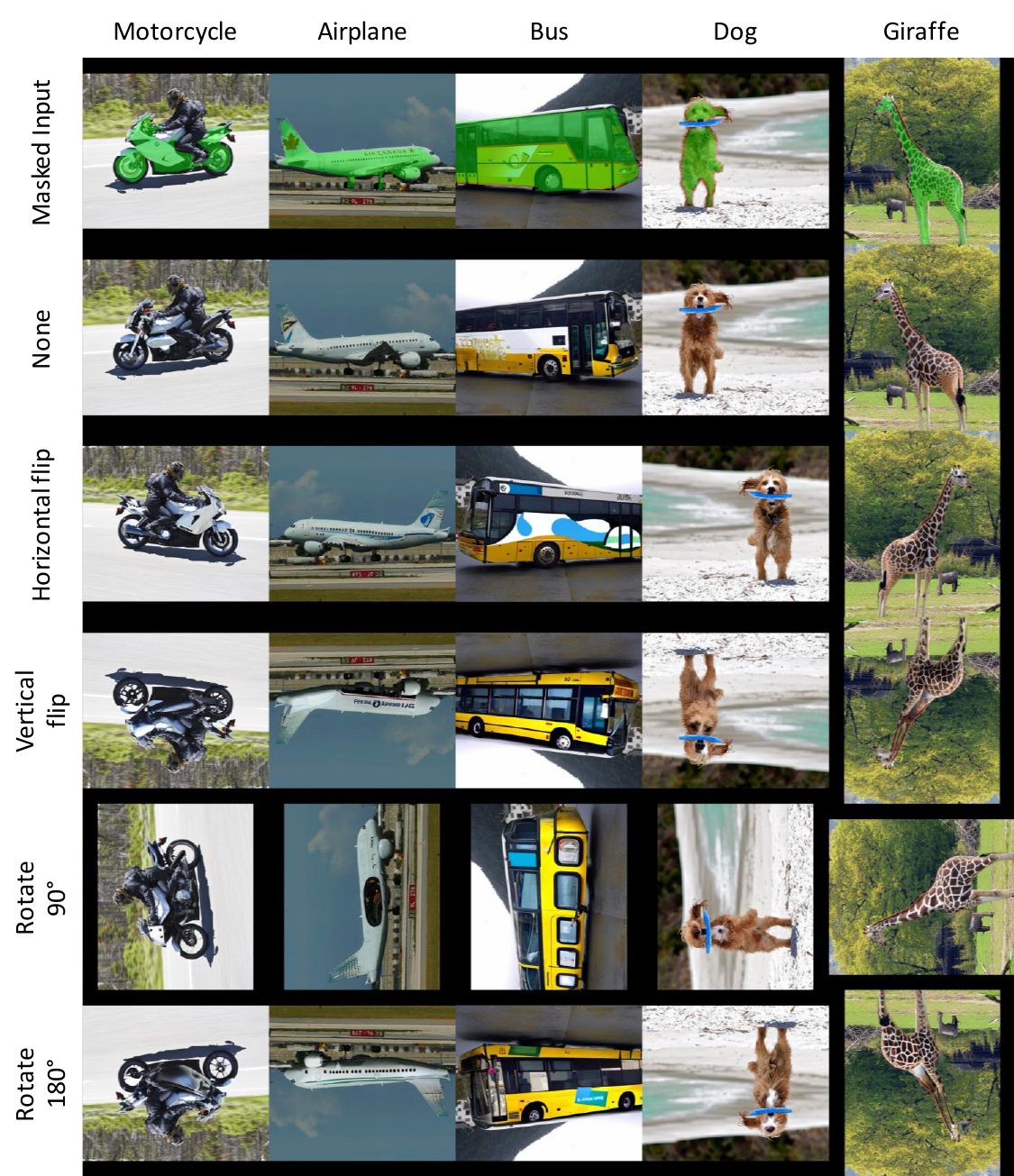}
\end{center}
   \caption{Flipping and rotating augmentations applied to input images. All images were padded if needed and then cropped, on the left side the additional augmentations applied to the input. On top the object class to inpaint. We used an erosion kernel of size 12 the standard prompt and no negative prompt. Masked input shown in first row.}
\label{fig:rotation}
\end{figure*}

\begin{figure*}
\begin{center}
\includegraphics[width=\textwidth]{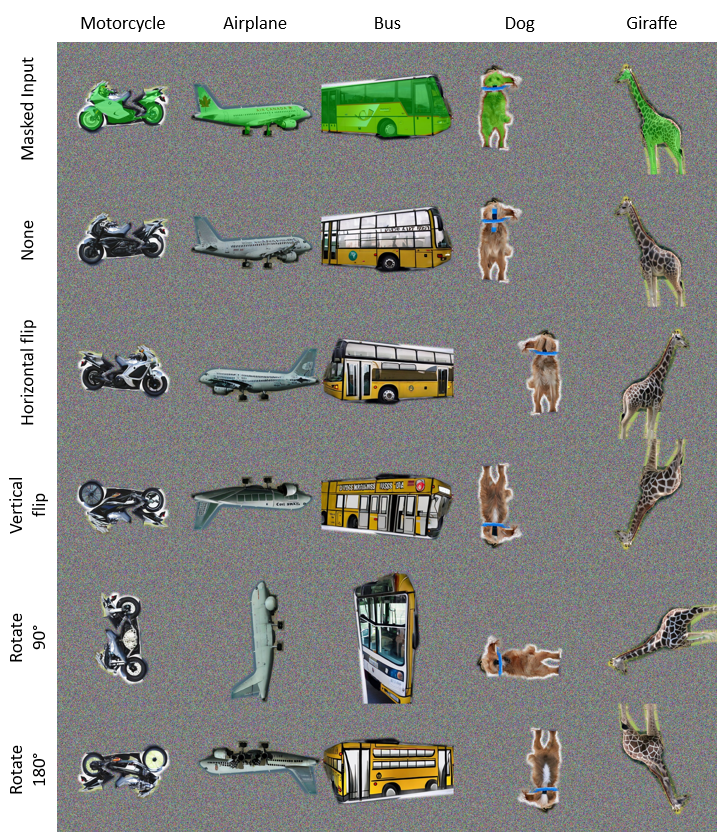}
\end{center}
   \caption{Flipping and rotating augmentations applied to input images where the background has been replaced with noise. All images were padded if needed and then cropped, on the left side the additional augmentations applied to the input. On top the object class to inpaint. We used an erosion kernel of size 12 the standard prompt and no negative prompt. Masked input shown in first row.}
\label{fig:rotation_noise}
\end{figure*}

\textbf{Flipping and Rotating}. As we can see in Figure \ref{fig:rotation}, several flipping and rotation augmentations have been applied to the input image. This results show that the method can follow horizontal flips but is not rotation invariant. Especially in the ``Bus'' example it is noticeable that it fails when we flip it vertical or rotate it by 180°. It tries to generate images where the tires face downwards even if the results become unnatural as seen when applying a vertical flip or a rotation. As seen in Figure \ref{fig:rotation_noise}, background information is not needed for flipping and rotating. In the ``Bus'' example it is still the case that the generation is biased towards facing downwards. We assume this biased is introduced in the training data.

\begin{figure*}
\begin{center}
\includegraphics[height=0.9\textheight]{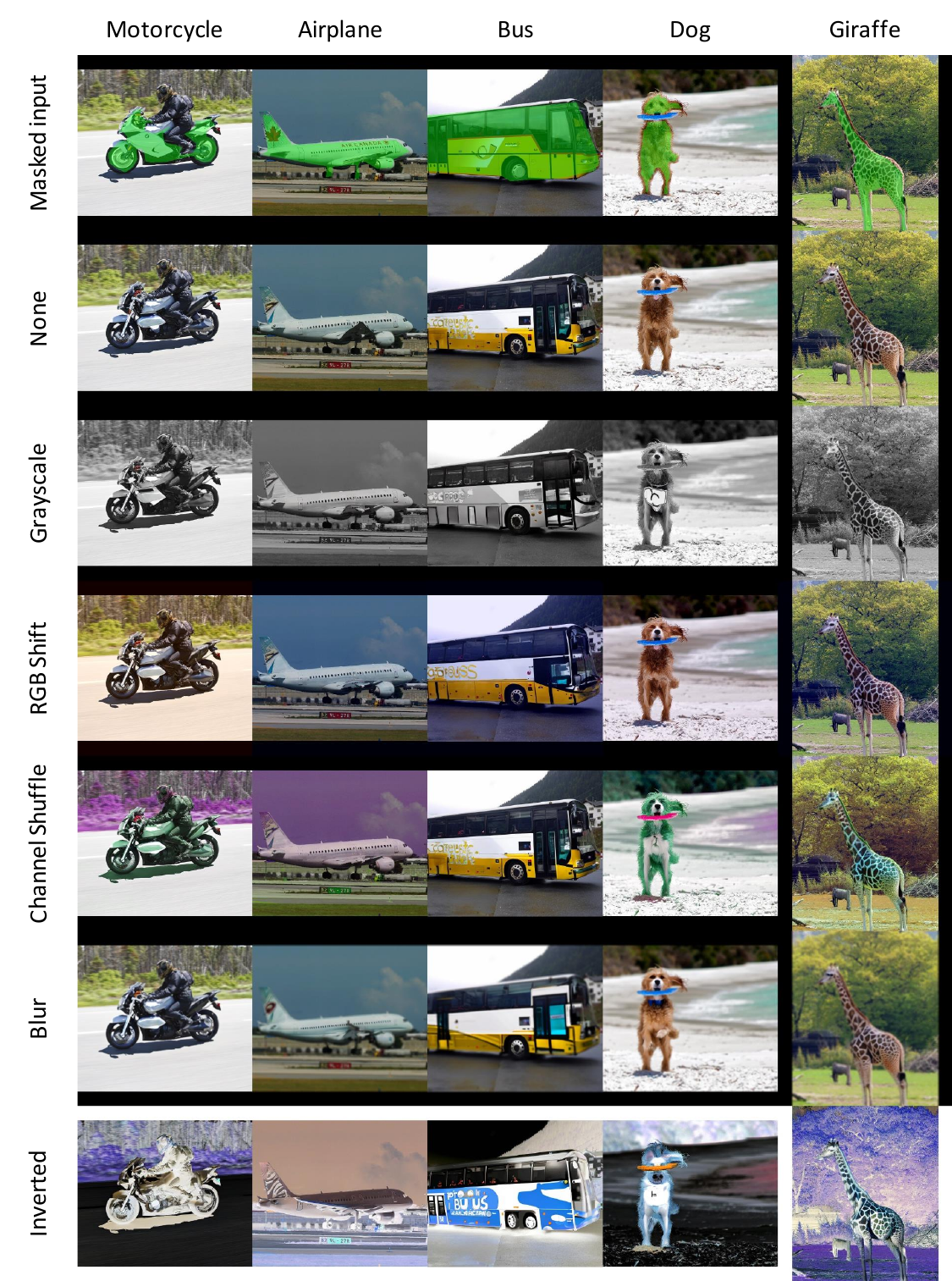}
\end{center}
   \caption{Pixel-level transforms applied to input images. All images were padded if needed and then cropped, on the left side the additional augmentations applied to the input. On top the object class to inpaint. We used an erosion kernel of size 12 the standard prompt and no negative prompt. Masked input shown in first row.}
\label{fig:aug_color}
\end{figure*}

\textbf{Pixel-level transforms}. In Figure \ref{fig:aug_color} we applied several pixel-level transforms on the input image. The method can follow pixel level transforms and apply them on the inpainted parts. Even when the colors seem unnatural for a specific object, as it is the case when we shuffle the channels or invert the values, it can inpaint defining visual features of the object and at the same time follow the pixel-level transform.

\section{Segmentation AP of Object Classes}
\begin{figure*}
\begin{center}
\includegraphics[width=\textwidth]{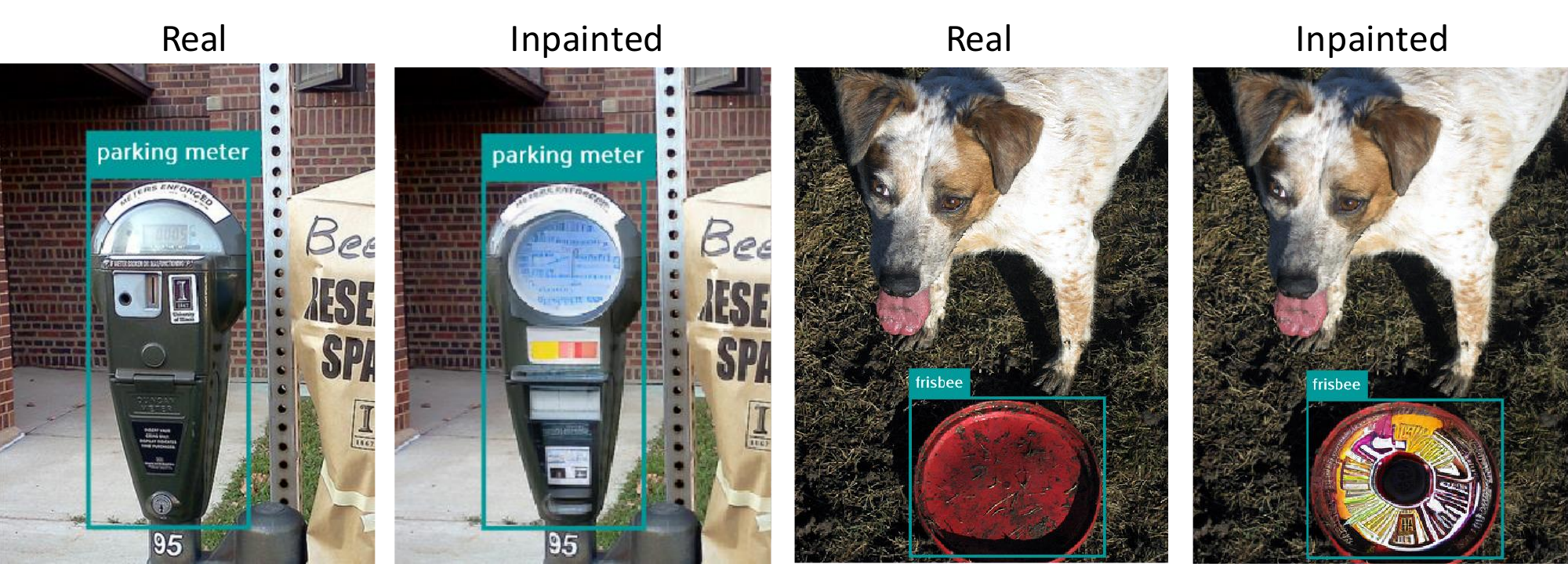}

\end{center}
   \caption{Two examples comparing real images and inpainted images of parking meter and frisbee.}
\label{fig:bad_classes}
\end{figure*}
Not all object classes get a performance increase when augmenting them with the proposed method. Notable are the COCO objects parking meter and firsbee that can be seen in Figure \ref{fig:bad_classes}. The parking meter contains fine details and text that the underlying model can not reproduce. With the frisbee it sometimes happens that strange color artifacts get inpainted as seen in the Figure \ref{fig:bad_classes}. This are things were our method has hardly any influence. From the perspective of the method, the goal was achieved of filling the whole object mask with related content. To improve the content one, may consider fine tuning the underlying model.

\section{Design Ideas}
\begin{figure*}
\begin{center}
\includegraphics[height=0.95\textheight]{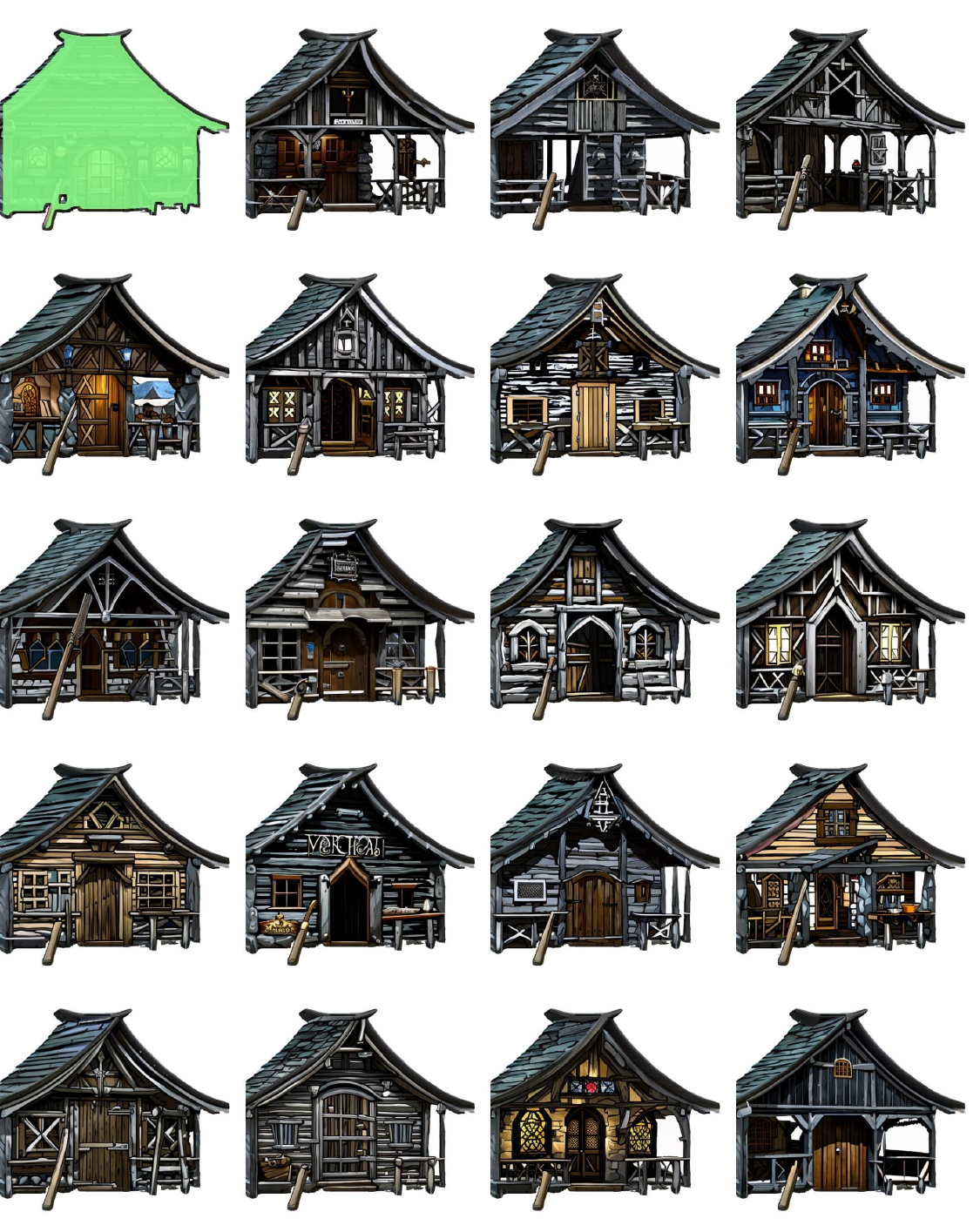}
\end{center}
   \caption{Digital art variations. Top left is the masked input. The other images are variations.}
\label{fig:art}
\end{figure*}

The proposed method is far more general than only augmenting an instance segmentation dataset. It is a good fit to get creative design ideas as shown in Figure \ref{fig:art}. The input is an image of a medieval tavern and a related mask (top left image in Figure \ref{fig:art}). Given the prompt ``illustration of a medieval tavern, high fantasy, epic, digital art'' and the mentioned negative prompt we can create various variations with ease. This can be applied to several fields such as digital design, architecture, and art.


{\small
\bibliographystyle{ieee_fullname}
\bibliography{egbib}
}